\documentclass{article}

\usepackage[preprint]{neurips_2026}

\usepackage[utf8]{inputenc} 
\usepackage[T1]{fontenc}    
\usepackage{hyperref}       
\usepackage{url}            
\usepackage{booktabs}       
\usepackage{amsfonts}       
\usepackage{amsmath}        
\usepackage{amssymb}        
\usepackage{nicefrac}       
\usepackage{microtype}      
\usepackage{xcolor}         
\usepackage{graphicx}       
\usepackage{capt-of}        
\usepackage{enumitem}       
\usepackage{wrapfig}        
\usepackage{tcolorbox}      
\tcbuselibrary{listings,breakable}
\newtcolorbox{promptbox}[1][]{
    colback=gray!5,
    colframe=gray!60,
    fonttitle=\bfseries\small,
    title=#1,
    breakable,
    left=4pt, right=4pt, top=2pt, bottom=2pt,
    boxrule=0.5pt,
    fontupper=\small\ttfamily
}

\title{ARES: Automated Rubric Synthesis for Scalable LLM Reinforcement Learning}

\author{%
  Xiaoyuan Li$^{1}$ \thanks{Equal contribution.}\footnotemark[1] \quad
  Keqin Bao$^{2}$ \footnotemark[1] \quad
  Moxin Li$^{3}$ \quad
  Yubo Ma$^{2}$ \quad \\
  \textbf{Yichang Zhang}$^{2}$ \quad 
  \textbf{Wenjie Wang}$^{1}$ \quad 
  \textbf{Fuli Feng}$^{1}$ \quad 
  \textbf{Dayiheng Liu}$^{2}$ \quad \\
  $^{1}$University of Science and Technology of China \quad
  $^{2}$Alibaba Group \quad \\
  $^{3}$National University of Singapore \\
}

\begin{document}

\maketitle
\begin{abstract}

Rubric-based rewards offer a promising way to extend reinforcement learning (RL) for large language models beyond tasks with automatically verifiable answers. 
However, scaling rubric-based RL remains challenging: existing approaches often rely on expert-written rubrics and manually constructed question sets, while fixed task-level rubrics may fail to capture the evaluation requirements of individual questions. 
We propose \textbf{ARES} (\textbf{A}utomated \textbf{R}ubric synth\textbf{E}sis for \textbf{S}calable RL), a framework for automatically constructing rubric-based RL data at scale. 
Starting from raw pretraining documents, ARES converts source knowledge into self-contained question-answer pairs and co-generates question-specific weighted rubrics, enabling instance-level reward supervision for open-ended responses. 
To improve diversity and quality, ARES conditions generation on domain labels and persona information, and applies validation filters for question self-containment, answer faithfulness, and rubric validity. 
Using ARES, we construct \textbf{100K} rubric-annotated instances across ten domains. 
Experiments on seven benchmarks show that rubric-based RL trained with ARES, outperforms continual pretraining, supervised fine-tuning, and binary-reward RL, with the largest gains on multi-dimensional open-ended tasks such as healthcare and instruction following.

\end{abstract}

\section{Introduction}

Large Language Models (LLMs) are typically developed through large-scale pre-training on web-scale text corpora, which equips them with broad linguistic patterns and knowledge~\citep{yang2024qwen2,together2023redpajama}, followed by post-training to better align them with human instructions and enhance their problem-solving abilities~\citep{ross2011reduction,bachmann2024pitfalls,shao2024deepseekmath}. Among recent post-training methods, Reinforcement Learning with Verifiable Rewards (RLVR) has become a particularly influential recipe for improving reasoning capabilities~\citep{deepseek2025r1,openai2024o1}. By optimizing models against automatically checkable outcomes, RLVR has achieved strong results in domains such as mathematics, coding, and short-answer reasoning. However, its success also exposes a fundamental limitation: \textit{scalable RL training is currently concentrated in tasks where correctness can be cheaply and reliably verified}.

This dependence on verifiable outcomes limits RLVR in two important ways. Firstly, such verifiable rewards are mainly available for tasks with automatically checkable final answers, such as math, coding, and short-answer tasks~\citep{deepscaler2025,openr1math2025,guha2025openthoughts, cen2025webscale}, limiting the diversity of domains and task types that can be reliably scaled. Secondly, the reward is sparse and often binary-like, making RL optimization difficult~\citep{lightman2023let}. Therefore, RLVR is powerful but narrow: it scales well only where verification is cheap, and provides limited supervision over how a response should satisfy complex, multi-dimensional requirements. 
Rubric-based rewards~\citep{gunjal2025rar} offer a promising alternative by evaluating responses against a set of weighted criteria, thereby decomposing response quality into multiple dimensions and providing finer-grained learning signals.

Despite their promise, scaling rubric-based RL remains challenging for two reasons. 
First, constructing rubric-annotated data usually requires experts to manually write questions, reference answers, and evaluation criteria, making it difficult to scale. Second, many rubrics are defined at the task or dataset level, so they provide general evaluation dimensions but cannot specify what should be emphasized for each individual question.
To address these issues, we aim to automatically synthesize large-scale rubric-based RL datasets with question-specific evaluation criteria and importance weights, enabling the reward to emphasize the aspects most critical to answer quality for each question.


In this work, we propose \textbf{ARES} (\textbf{A}utomated \textbf{R}ubric synth\textbf{E}sis for \textbf{S}calable RL), a framework for automatically constructing large-scale rubric-based RL training data from raw pretraining documents. Given a document containing knowledge to be transferred into the RL stage, ARES generates a self-contained question-answer pair that can be answered without access to the source document, together with a rubric consisting of question-specific evaluation criteria and their importance weights. To further improve data diversity, ARES conditions the generation process on domain and persona signals, producing questions and response styles that cover diverse scenarios. Using this pipeline, we synthesize \textbf{100K} rubric-annotated instances across ten domains, enabling RL training over a much broader range of open-ended capabilities than conventional verifiable-reward settings can support. 
Our main contributions are as follows:

\begin{itemize}[leftmargin=*]
\item We propose ARES, a scalable document-to-RL pipeline that automatically transforms raw pretraining documents into RL instances with self-contained questions, reference answers, and question-specific weighted rubrics.

\item We construct a large-scale rubric-annotated dataset of \textbf{100K} examples spanning ten domains, where each instance provides fine-grained reward supervision.

\item We evaluate ARES across seven diverse benchmarks, showing that it outperforms continual pretraining, supervised fine-tuning, and binary-reward RL, with especially large gains on open-ended tasks such as healthcare (+6.4) and instruction following (+15.5).

\end{itemize}

\section{Related Work}

\paragraph{Data Synthesis for RL.}
Early efforts focused on curating domain-specific datasets for mathematics and code~\citep{deepscaler2025,openr1math2025}. Subsequent work expanded coverage to broader web-scale sources~\citep{yuan2025naturalreasoning,guha2025openthoughts,bercovich2025nemotron}, increasing diversity while remaining focused on tasks with verifiable ground truth. Our work most closely relates to \textsc{Webscale-RL}~\citep{cen2025webscale}, which grounds both questions and answers in pretraining documents to enable large-scale RL data synthesis. However, all prior approaches generate short verifiable answers and rely on binary reward signals, restricting RL to tasks with unambiguous ground truth. We extend this paradigm to open-ended domains by synthesizing multi-dimensional rubric alongside QA pairs.

\paragraph{Reward Design for RL Training.}
The dominant approach uses binary rewards based on exact string matching~\citep{cen2025webscale,deepscaler2025,openr1math2025}, which is robust and stable but limited to verifiable tasks. An alternative is using LLMs as judges~\citep{zheng2023judging}, where a model evaluates response quality holistically; this is flexible but can be inconsistent and expensive. Process reward models~\citep{lightman2023let,wang2024math} provide step-level feedback for multi-step reasoning, improving training signal density over outcome-level binary rewards, but require substantial annotation to train. Rubric-based evaluation, common in educational assessment and long-form writing evaluation~\citep{writingbench2025}, decomposes quality into multiple weighted dimensions, enabling structured, transparent, and reproducible assessment. \citet{gunjal2025rar} recently demonstrate the promise of rubric-based rewards for RL, showing gains over binary and Likert-scale baselines on medical and science domains. However, their approach relies on curated question sets from specific domains and requires separate rubric generation passes. We extend this direction by co-generating rubrics with QA pairs from raw pretraining documents in a single inference pass, enabling rubric-based RL at an order of magnitude greater scale and domain diversity.

\paragraph{Pretraining-Stage and Large-Scale RL.}
Recent work explores applying RL at earlier training stages or at larger scale. \citet{cen2025webscale} and \citet{li2025rlpretraining} demonstrate that RL on pretraining-scale data yields significant gains. \citet{liu2025prorl,liu2025scalingrl} investigate prolonged RL training and its scaling properties. Our work complements these efforts by focusing on the reward signal design rather than the scale of training, showing that richer rewards enable RL to benefit a broader set of capabilities including open-ended domains.

\begin{figure}[t]
    \centering
    \includegraphics[width=0.8\linewidth]{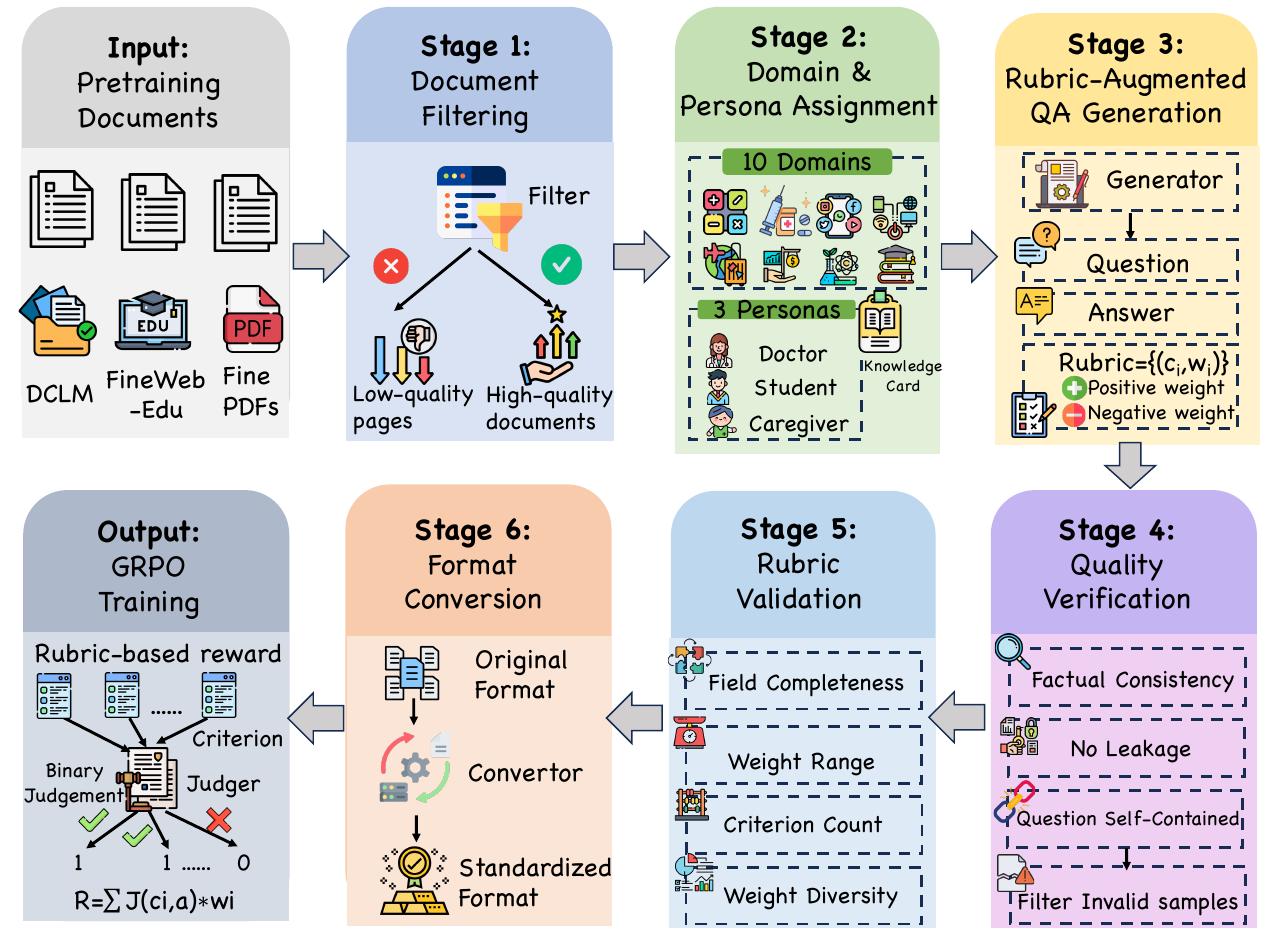}
    \caption{Overview of the six-stage ARES pipeline. Starting from raw pretraining documents, ARES performs document filtering, domain and persona conditioning, rubric-augmented QA generation, quality verification, rubric validation, and format conversion to produce training instances.}
    \label{fig:pipeline}
\end{figure}

\section{Methodology}

\subsection{Preliminaries}

\paragraph{RLVR.}
RLVR is a post-training paradigm that improves LLMs by optimizing them against automatically checkable outcomes. 
Given a question $q$, a policy samples multiple candidate responses, each of which receives a reward from an external verifier according to answer correctness or task success. 
RLVR has been particularly effective for domains such as mathematics and coding, where final answers or program outputs can be reliably verified.

In this work, we use Group Relative Policy Optimization (\textbf{GRPO})~\citep{shao2024deepseekmath} as the RL optimizer. 
For each question $q$, the behavior policy $\pi_{\theta_{\mathrm{old}}}$ samples a group of $G$ responses $\{y_i\}_{i=1}^{G}$, and each response $y_i$ is assigned a reward $R_i$. 
GRPO optimizes the policy by comparing responses within the same group:
\begin{equation}
    \mathcal{J}_\text{GRPO}(\theta)
    =
    \mathbb{E}_{q,\{y_i\}_{i=1}^{G}}
    \left[
    \frac{1}{G}\sum_{i=1}^{G}
    \min\left(
        \rho_i \hat{A}_i,
        \mathrm{clip}(\rho_i,1-\varepsilon,1+\varepsilon)\hat{A}_i
    \right)
    -
    \beta\,\mathrm{KL}(\pi_\theta\|\pi_\text{ref})
    \right],
    \label{eq:grpo}
\end{equation}
where
\begin{equation}
    \rho_i =
    \frac{\pi_\theta(y_i \mid q)}
         {\pi_{\theta_{\mathrm{old}}}(y_i \mid q)},
    \qquad
    \hat{A}_i =
    \frac{R_i - \operatorname{mean}(\{R_j\}_{j=1}^{G})}
         {\operatorname{std}(\{R_j\}_{j=1}^{G})+\epsilon_A}.
\end{equation}
Here, $\rho_i$ is the importance ratio, $\hat{A}_i$ is the group-normalized advantage, $\varepsilon$ is the clipping threshold, $\beta$ controls KL regularization, $\pi_\text{ref}$ is the reference policy, and $\epsilon_A$ is a small constant for numerical stability. 
In standard RLVR, $R_i$ is usually provided by a verifier that checks outcome-level correctness. 
While effective, such rewards are naturally restricted to tasks with reliable automatic verification and often provide sparse supervision.

\paragraph{RL with Rubric-Based Rewards.}
Rubric-based rewards extend RL training to more open-ended tasks by replacing outcome-only verification with multi-dimensional response evaluation~\citep{gunjal2025rar}. 
Instead of judging a response solely by final-answer correctness, a rubric specifies a set of criteria that describe different aspects of response quality, such as factual correctness, completeness, reasoning consistency, instruction following, or avoidance of unsupported claims.

Formally, for a question $q$, a rubric is defined as
\begin{equation}
    \mathcal{R}(q)=\{(c_k,w_k)\}_{k=1}^{N},
\end{equation}
where $c_k$ denotes the $k$-th evaluation criterion and $w_k$ denotes its importance weight. 
Given a candidate response $y$, an LLM judge $J_\phi$ scores whether $y$ satisfies each criterion. 
The rubric-based reward is computed as
\begin{equation}
    R_\text{rubric}(q,y;\mathcal{R})
    =
    \sum_{k=1}^{N} w_k \cdot J_\phi(q,y,c_k),
    \label{eq:rubric_reward}
\end{equation}
where $J_\phi(q,y,c_k)\in[0,1]$ measures the degree to which response $y$ satisfies criterion $c_k$. 
``1'' encourages desirable properties, while negative weights can penalize common failure modes, such as hallucination, irrelevant information, or unsafe recommendations. 
During RL training, the rubric reward is used as the response reward in Eq.~\eqref{eq:grpo}, i.e.,
\begin{equation}
    R_i = R_\text{rubric}(q,y_i;\mathcal{R}).
\end{equation}

Compared with final-answer verification, rubric-based rewards provide denser and more diagnostic supervision because they identify which aspects of a response are satisfactory or deficient. 
However, existing rubric-based RL methods typically rely on manually designed criteria, expert-written rubrics, or fixed task-level evaluation dimensions~\citep{gallego2025configurable,akyurek2025prbench}. 
This limits both scalability and granularity: manually written rubrics are costly to construct at scale, while fixed rubrics may fail to capture the specific evaluation requirements of individual questions. 
ARES addresses this bottleneck by automatically synthesizing question-specific weighted rubrics together with self-contained question-answer pairs from raw pretraining documents.

\subsection{Data Pipeline}


To tackle these limitations, ARES constructs scalable rubric-based RL data by converting raw pretraining documents into instances with question-specific weighted rubrics. 
Given a document $d$, the pipeline produces a tuple
\begin{equation}
    d \xrightarrow{\text{ARES}} (q, a^*, \mathcal{R}_q),
\end{equation}
where $q$ is a self-contained question, $a^*$ is the reference answer, and 
$\mathcal{R}_q=\{(c_k^{(q)},w_k^{(q)})\}_{k=1}^{N_q}$ is a rubric tailored to $q$. 
During RL training, $\mathcal{R}_q$ is used to compute the reward for each sampled response:
\begin{equation}
     R_\text{rubric}(q,y;\mathcal{R}_q)
    =
    \sum_{k=1}^{N_q} w_k^{(q)} \cdot J(q,y,c_k^{(q)}).
    \label{eq:reward}
\end{equation}
The pipeline consists of three major phases: source preparation, rubric-augmented QA co-generation, and quality control.

\paragraph{Source preparation.}
ARES first filters raw pretraining documents using an LLM-based classifier, removing documents that are low-quality, boilerplate-heavy, or lack sufficient context for question generation. 
To enhance the diversity of both questions and rubrics, we further annotate each retained document with a semantic domain and three target personas. For instance, the same medical article may be used to generate questions tailored to medical professional, psychology student, or family caregiver, resulting in different levels of background knowledge, answer styles, and reasoning depth.


\paragraph{Rubric-augmented QA co-generation.}
Given a filtered document together with its domain and persona conditioning, ARES uses an LLM generator $M_\text{gen}$\footnote{Document filtering and rubric-augmented QA generation use Gemini-3.1-Pro-Preview; domain/persona conditioning, question and rubric quality control use Claude-Sonnet-4.5.} to jointly produce the question, reference answer, and question-specific rubric:
\begin{equation}
    (q, a^*, \mathcal{R}_q) = M_\text{gen}(d, \delta, p).
\end{equation}
The generated question is required to be self-contained, so that it can be answered without access to the source document. 
The reference answer is grounded in the document, while the rubric specifies how candidate responses should be evaluated for this particular question. 
By co-generating $q$, $a^*$, and $\mathcal{R}_q$ in a single pass, ARES aligns the evaluation criteria with both the source knowledge and the intended answer, rather than relying on generic task-level rubrics. 
The rubric contains weighted positive criteria for desired response properties and negative criteria for common failure modes, with the number of criteria adapted to question complexity.


\paragraph{Quality control.}
ARES applies validation before using an instance for training. 
QA validation checks whether the question is self-contained, whether the reference answer is faithful to the source document, and whether the answer cannot be trivially inferred from surface cues in the question alone. 
Rubric validation checks whether each criterion is well-formed, whether the rubric covers the key requirements of the question, and whether the weights are valid and sufficiently diverse. 
Instances that fail validation are discarded. 
The remaining tuples are serialized with metadata such as domain, persona, and instance identifiers to form GRPO-compatible training data. Finally, we perform data deduplication and remove overlaps in the evaluation sets.

\begin{table}[t]
    \centering
    \caption{Comparison of existing datasets and RL training pipelines across four key desiderata. \checkmark~indicates the property is satisfied; $\times$ indicates it is not; /~indicates the property is not applicable. \textbf{Multi-Domain}: covers diverse domains beyond math and code. \textbf{Rubric Rewards}: provides multi-dimensional rubric-based reward signals rather than binary verification. \textbf{Doc-Grounded}: questions and answers are grounded in source pretraining documents. \textbf{Scalable}: the pipeline can be readily scaled to larger data without manual curation. ARES is the only approach satisfying all four criteria.}
    \label{tab:comparison}
    \small
    \begin{tabular}{lcccc}
    \toprule
    \textbf{Dataset} & \textbf{Multi-Domain} & \textbf{Rubric Rewards} & \textbf{Doc-Grounded} & \textbf{Scalable} \\
    \midrule
    DeepScaleR~\citep{deepscaler2025}           & $\times$ & $\times$ & $\times$ & $\times$ \\
    OpenR1-Math~\citep{openr1math2025}          & $\times$ & $\times$ & $\times$ & $\times$ \\
    OpenThoughts3~\citep{guha2025openthoughts}  & $\times$ & $\times$ & $\times$ & $\times$ \\
    Nemotron~\citep{bercovich2025nemotron}      & $\times$ & $\times$ & $\times$ & $\times$ \\
    RaR~\citep{gunjal2025rar}                         & $\times$   & \checkmark & $\times$ & $\times$ \\
    NaturalReasoning~\citep{yuan2025naturalreasoning} & \checkmark & $\times$ & \checkmark &  \checkmark \\
    Webscale-RL~\citep{cen2025webscale}               & \checkmark & $\times$ & \checkmark &  \checkmark \\
    \midrule
    \textbf{ARES (ours)} & \checkmark & \checkmark & \checkmark & \checkmark \\
    \bottomrule
    \end{tabular}
    \end{table}

    

\begin{figure*}[t]
    \centering
    \begin{minipage}[t]{0.48\textwidth}
        \vspace{0pt}
        \centering
        \includegraphics[height=0.18\textheight]{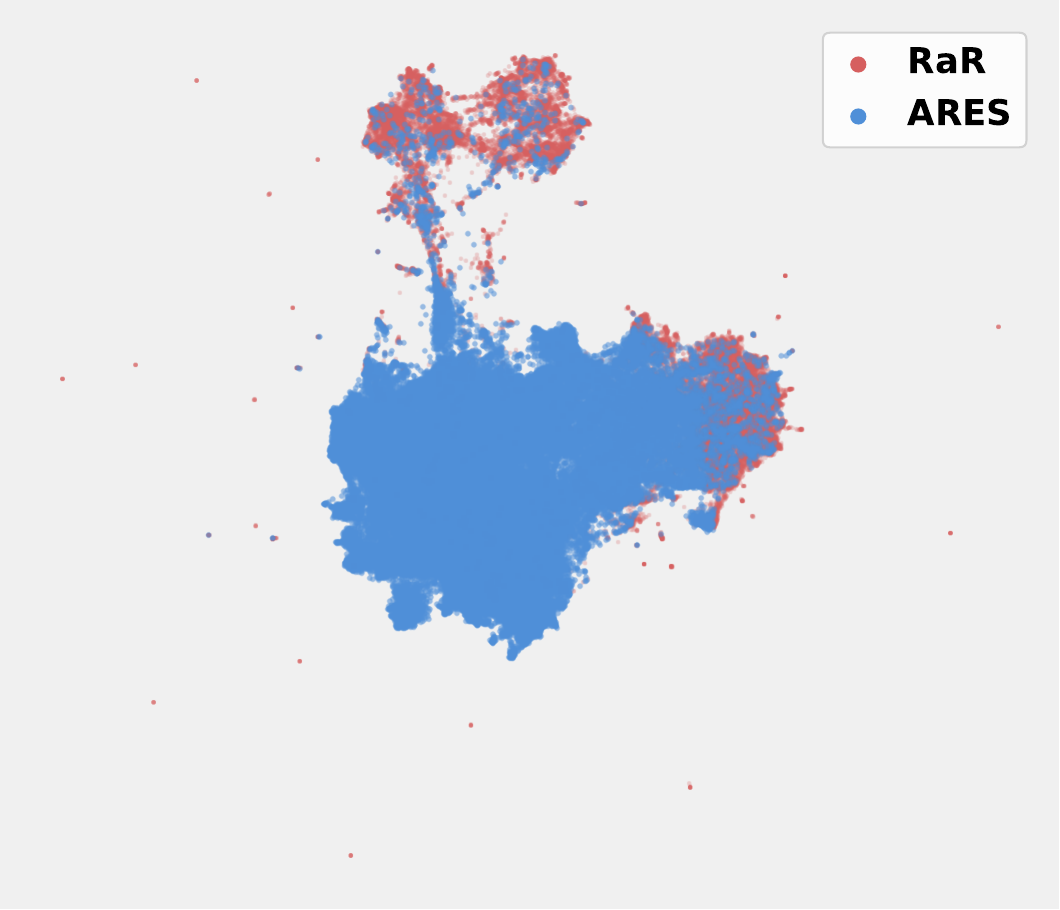}
        \captionof{figure}{UMAP visualization of questions for RaR and ARES by Qwen3-Embeddings-0.6B.}
        \label{fig:domain_dist}
    \end{minipage}\hfill
    \begin{minipage}[t]{0.48\textwidth}
        \vspace{0pt}
        \centering
        \includegraphics[height=0.18\textheight]{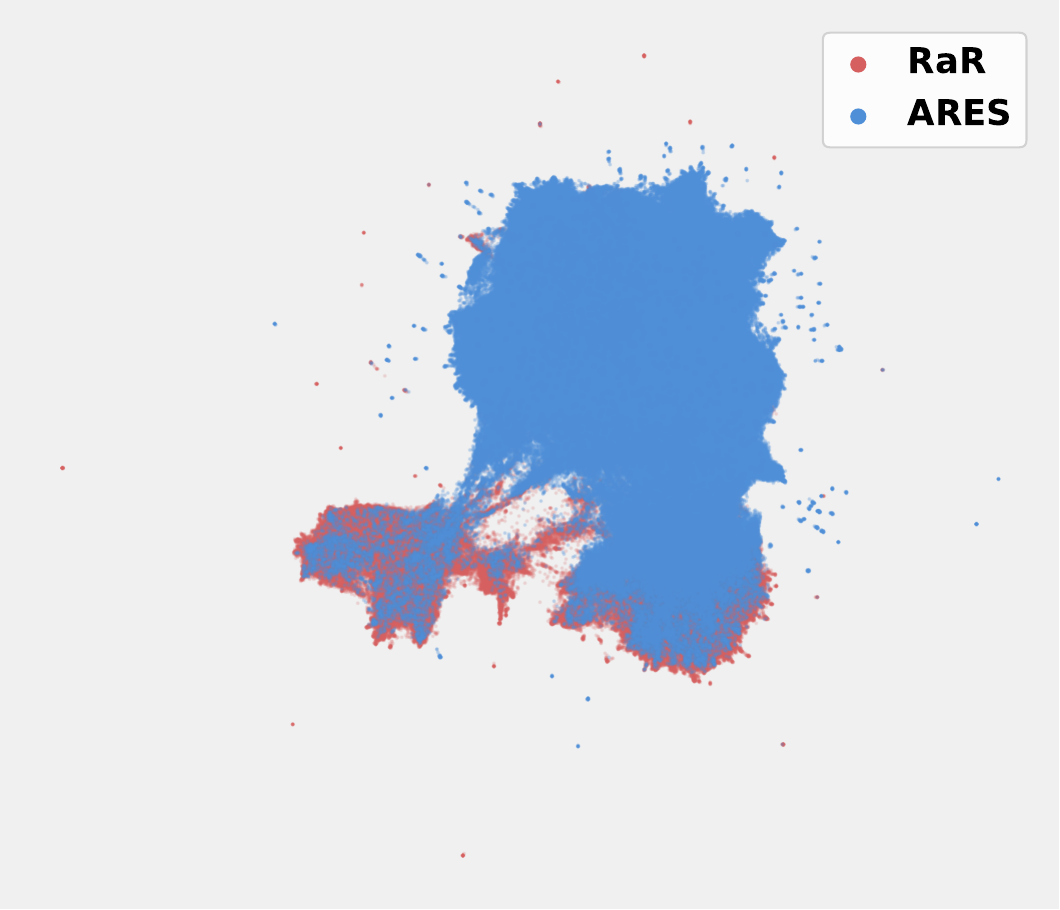}
        \captionof{figure}{Distribution of rubric criteria in RaR and ARES by Qwen3-Embeddings-0.6B.}
        \label{fig:rubric_dist}
    \end{minipage}
    \vspace{-20pt}
\end{figure*}

\subsection{Comparison with Prior Work}

Table~\ref{tab:comparison} positions ARES relative to existing datasets and pipelines along four key dimensions. Math- and code-focused RL datasets~\citep{deepscaler2025,openr1math2025,guha2025openthoughts,bercovich2025nemotron} provide verifiable binary rewards but collect queries from limited curated sources, restricting both domain coverage and scalability. Webscale-RL~\citep{cen2025webscale} and NaturalReasoning~\citep{yuan2025naturalreasoning} extend data synthesis to diverse, document-grounded sources, but remain limited to binary reward signals, excluding open-ended domains where exact-match verification is infeasible. RaR~\citep{gunjal2025rar} introduces rubric-based rewards but relies on curated question sets from science and medicine domains, limiting both scale and coverage. As visualized in Figure~\ref{fig:domain_dist} and Figure~\ref{fig:rubric_dist}, RaR (red) is confined to two narrow clusters, whereas ARES (blue) not only fully covers those regions but extends to substantially broader areas. ARES is the first pipeline to simultaneously satisfy all four criteria: it draws from large-scale pretraining corpora, covers ten diverse domains, grounds all outputs in source documents, and produces multi-dimensional rubric rewards---enabling RL training at a breadth and scale previously unattainable.

\begin{wraptable}{l}{0.48\textwidth}
    \centering
    \small
    \vspace{-20pt}
    \caption{Domain distribution of the ARES dataset spanning ten domains.}
    \label{tab:domain_dist}
    \begin{tabular}{lr}
    \toprule
    Domain & \# Examples \\
    \midrule
    Social Science              & 18,878 \\
    Technology \& Engineering   & 18,321 \\
    Medicine \& Health          & 13,974 \\
    Travel \& Lifestyle         & 12,767 \\
    Commerce \& Economics       & 12,439 \\
    Natural Science             & 10,154 \\
    Education                   &  8,822 \\
    Other                       &  3,931 \\
    Coding                      &  1,631 \\
    Math                        &    930 \\
    \midrule
    \textbf{Total}              & \textbf{101,847} \\
    \bottomrule
    \end{tabular}
    \vspace{2pt}
    \caption{Rubric statistics of the ARES dataset.}
    \label{tab:rubric_stats}
    \begin{tabular}{lr}
    \toprule
    Metric & Value \\
    \midrule
    \#Criteria         & 1,108,163 \\
    Avg. Criteria      & 10.88 \\
    Median Criteria    & 11 \\
    Positive Criteria  & 817,047 \\
    Negative Criteria  & 291,116 \\
    \bottomrule
    \end{tabular}
    \vspace{-10pt}
\end{wraptable}

\subsection{Dataset Statistics}

We run ARES on a small subset of pretraining corpora from DCLM~\citep{li2024datacomp}, FineWeb-Edu~\citep{penedo2024fineweb}, and FinePDFs~\citep{finepdfs2024}. 
Specifically, we first sample and filter approximately 0.1B tokens from these corpora as the source pool for rubric-augmented QA synthesis. 
After document filtering, co-generation, and validation, ARES produces 101,847 rubric-annotated QA instances, corresponding to an overall retention rate of 71.6\%. 

As summarized in Table~\ref{tab:domain_dist}, the resulting dataset spans ten domains, with Social Science and Technology \& Engineering constituting the two largest categories. 
Coding and Math account for smaller proportions in the generated dataset, reflecting the domain composition after document filtering and classification. 
The remaining domains cover medicine, lifestyle, commerce, natural science, education, and other topics. 

Table~\ref{tab:rubric_stats} reports the resulting rubric statistics. 
The dataset contains 1,108,163 criteria in total, with an average of 10.88 criteria per instance and a median of 11, consistent with our complexity-aware generation design. 
Among them, 817,047 are positive criteria that reward desired response properties, while 291,116 are negative criteria that penalize common failure modes. 
This mixture is intended to provide both positive incentives and discriminative penalties during rubric-based RL training.
\section{Experiments}

\subsection{Experimental Setup}

\paragraph{Base Model.}
All experiments use Qwen3-4B-Base~\citep{yang2024qwen2} as the foundation model, which we train from scratch in each setting without any instruction tuning warmup unless specified.

\paragraph{Baselines.}
We compare against four baselines trained from the same foundation model. 
(1) \textbf{CPT} continues pretraining on the same source documents used by ARES. 
(2) \textbf{Natural Reasoning} fine-tunes on the NaturalReasoning dataset~\citep{yuan2025naturalreasoning}, serving as a strong off-the-shelf SFT baseline. 
(3) \textbf{Webscale} applies GRPO with binary rewards on reformatted QA pairs, following \textsc{Webscale-RL}~\citep{cen2025webscale}, to isolate the effect of rubric-based rewards. 
(4) \textbf{ARES-SFT} fine-tunes on ARES-generated QA pairs using reference answers as targets, isolating the effect of rubric-based RL from the ARES data construction.


\paragraph{ARES-RL.}
We apply GRPO with rubric-based rewards (Equation~\ref{eq:reward}) following~\citet{zhou2025breaking}. We train for 3 epochs with batch size 128, using VeRL~\citep{sheng2025hybridflow} as the training backend and following the hyper-parameters from RuscaRL~\cite{zhou2025breaking}. For the rubric judge $J$ in Equation~\ref{eq:reward}, we use Qwen3-32B~\citep{yang2024qwen2} following~\citet{zhou2025breaking}.

\paragraph{Benchmarks.}
We evaluate on seven benchmarks covering knowledge reasoning, math, code generation, healthcare QA, writing quality, and instruction following: {MMLU-Pro}~\citep{wang2024mmlu}, {GSM8K}~\citep{cobbe2021gsm8k}, {HumanEval+}~\citep{chen2021humaneval}, {MBPP+}~\citep{austin2021mbpp}, {HealthBench}~\citep{healthbench2025}, {WritingBench}~\citep{writingbench2025}, and {IFEval}~\citep{zhou2023ifeval}.

\subsection{Main Results}

\begin{table}[t]
    \centering
    \caption{Main results on seven diverse benchmarks. \textbf{Bold} indicates the best result per column.}
    \label{tab:main}
    \small
    \resizebox{\columnwidth}{!}{%
    \begin{tabular}{lcccccccr}
    \toprule
    Method & MMLU-Pro & GSM8K & HumanEval+ & MBPP+ & HealthBench & WritingBench & IFEval & Avg \\
    \midrule
    CPT               & 46.02 & 82.34 & 32.32 & 59.40 & 35.04 & 36.98 & 39.39 & 47.36 \\
    NaturalReasoning & 47.96 & 81.50 & 32.93 & 61.15 & 32.94 & 36.77 & 28.15 & 45.91 \\
    Webscale          & 49.50 & 84.91 & 33.54 & 61.40 & 36.08 & 37.09 & 35.61 & 48.30 \\
    ARES-SFT          & \textbf{50.56} & 85.67 & 31.10 & 61.40 & 35.78 & 37.05 & 46.41 & 49.71 \\
    \midrule
    ARES-RL (ours)    & 49.36 & \textbf{86.96} & \textbf{34.76} & \textbf{63.16} & \textbf{41.45} & \textbf{38.24} & \textbf{54.88} & \textbf{52.69} \\
    \bottomrule
    \end{tabular}
    }
    \vspace{-2em}
    \end{table}
Table~\ref{tab:main} presents the main experimental results. We highlight several key findings:

\vspace{-5pt}
\paragraph{ARES-RL Achieves the Best Average Performance.}
ARES-RL achieves the strongest average performance among the compared training methods, surpassing CPT by 5.33 points, SFT-baseline (NaturalReasoning) by 6.78 points, and binary-reward RL (Webscale) by 4.39 points. 
Notably, ARES-RL also outperforms ARES-SFT by 2.98 points while using the same ARES-generated data. 
This comparison isolates the benefit of the training objective: SFT mainly learns to imitate reference answers, whereas rubric-based RL optimizes candidate responses against multi-dimensional evaluation criteria. 
As a result, the policy can improve aspects of response quality that are not fully captured by a single reference answer. 
These gains suggest that rubric-based rewards provide richer learning signals than imitation-based training or binary outcome verification.

\vspace{-5pt}
\paragraph{Performance on Multi-Dimensional Evaluation Benchmarks.}
The largest improvements appear on benchmarks where response quality depends on multiple evaluation dimensions. 
Compared with binary-reward RL, ARES-RL improves HealthBench from 36.08 to 41.45 (+5.37), IFEval from 35.61 to 54.88 (+19.27), and WritingBench from 37.09 to 38.24 (+1.15). 
These tasks are difficult to capture with a single binary outcome reward: healthcare responses require factuality, safety, and reasoning quality; instruction following depends on satisfying multiple constraints; and writing quality involves coherence, relevance, and style. 
Rubric-based rewards, as a family of reward designs, are therefore well suited to these settings because they decompose response quality into explicit criteria.

\vspace{-5pt}
\paragraph{Performance on Verifiable Reasoning and Coding Tasks.}
Beyond multi-dimensional evaluation benchmarks, ARES-RL also performs strongly on automatically scored reasoning and coding tasks. 
It achieves the best GSM8K score (86.96, +2.05 over Webscale) and leads on both code generation benchmarks: HumanEval+ (34.76, +1.22 over Webscale) and MBPP+ (63.16, +1.76 over Webscale). 
This is notable because Math and Coding are the two smallest domains in ARES, containing only 930 and 1,631 examples, respectively (Table~\ref{tab:domain_dist}). 
Despite limited domain-specific data, ARES-RL achieves the best results on all three corresponding benchmarks, suggesting that the benefits of rubric-based RL over diverse domains can transfer to structured reasoning and coding tasks.
The main exception is MMLU-Pro, where ARES-RL (49.36) is slightly below ARES-SFT (50.56). 
This may be because MMLU-Pro is a multiple-choice benchmark, for which supervised learning or exact-answer-oriented training provides a more direct signal than open-ended rubric-based optimization.

\subsection{Analysis}

\paragraph{Why Do Rubric Rewards Help Open-Ended Domains?}
The gains on HealthBench, IFEval, and WritingBench can be attributed to two factors. 
First, these benchmarks evaluate response quality along multiple dimensions that are difficult to capture with a single binary outcome reward. 
For example, a health response may be partially correct, context-dependent, or appropriately cautious; a rubric can separately evaluate medical accuracy, completeness of risk disclosure, treatment reasoning, and clarity of explanation. 
Similarly, instruction-following tasks often contain multiple constraints, such as format, length, keywords, language, and structural requirements, which can be naturally decomposed into individual rubric criteria. 
Second, rubric rewards provide a denser and more diagnostic reward signal for GRPO. 
A response that correctly identifies a diagnosis but inadequately explains treatment options can still receive partial credit for its correct components while being penalized for missing dimensions, guiding the policy toward more complete and reliable responses.

\begin{figure*}[t]
    \centering
    \includegraphics[width=\textwidth]{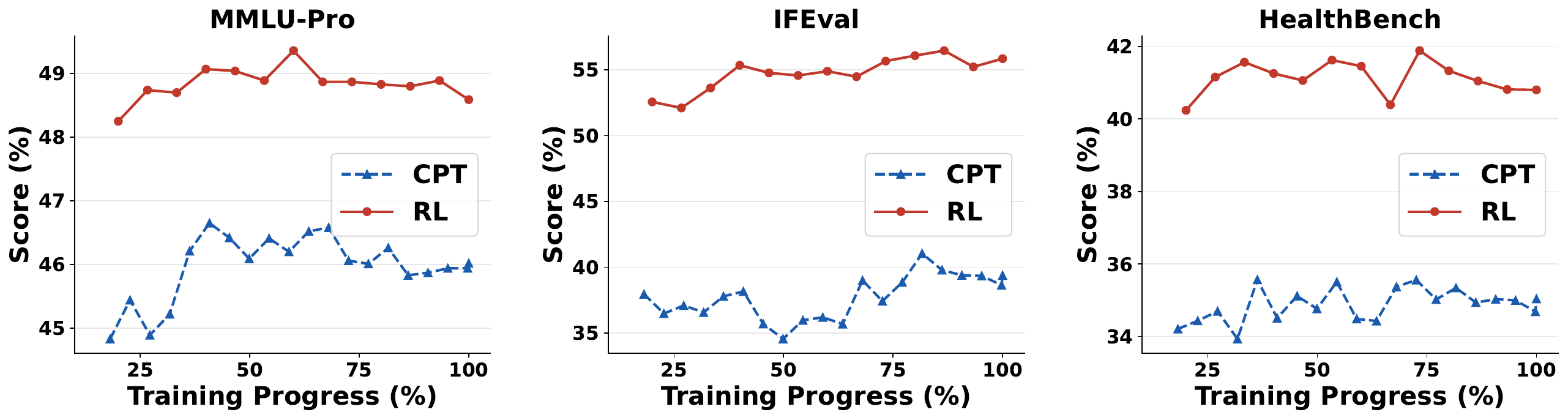}
    \vspace{-1em}
    \caption{Per-benchmark comparison between CPT and ARES-RL. }
    \vspace{-0.5em}
    \label{fig:cpt_vs_rl_all_benchmarks}
\end{figure*}

\vspace{-10pt}
\paragraph{From Passive Token Prediction to Targeted Reward Optimization.}
Figure~\ref{fig:cpt_vs_rl_all_benchmarks} compares CPT and ARES-RL using the same source-document pool. 
CPT learns from raw documents through next-token prediction, whereas ARES-RL converts them into QA-rubric instances and optimizes the policy with rubric-based rewards. 
The gap is modest on knowledge-intensive or automatically scored benchmarks, but becomes much larger on open-ended tasks, where ARES-RL improves over CPT by +6.41 on HealthBench and +15.49 on IFEval. 
This suggests that targeted rubric supervision is more effective than token-level prediction for capabilities requiring nuanced judgment, constraint satisfaction, and multi-dimensional response quality. 
By assigning credit and penalties across multiple criteria, ARES-RL turns raw documents into more focused post-training supervision.

\subsection{Ablation Study}

To understand the contribution of question-specific rubric design, we compare ARES-RL against three alternative reward strategies, all using the same training dataset and GRPO setup:(1) \textbf{Blind Judge}: an LLM judge evaluates response quality holistically without any rubric or reference answer, producing a direct scalar reward. (2) \textbf{General Rubric}: a fixed set of generic rubric criteria (e.g., \emph{accuracy}, \emph{completeness}, \emph{clarity}) is applied uniformly to all questions, rather than question-specific criteria. (3) \textbf{Reference Answer}: the judge evaluates by comparing the response against the reference answer, assigning a similarity-based score.
\begin{table}[t]
    \centering
    \caption{Ablation study on reward design. All variants use the same training dataset and GRPO setup. \textbf{Bold} indicates the best result per column.}
    \label{tab:ablation}
    \small
    \resizebox{\columnwidth}{!}{%
    \begin{tabular}{lcccccccr}
    \toprule
    Reward Strategy & MMLU-Pro & GSM8K & HumanEval+ & MBPP+ & HealthBench & WritingBench & IFEval & Avg \\
    \midrule
    Blind Judge      & 47.56 & 85.97 & 33.54 & 62.16 & 44.43 & 38.84 & 34.20 & 49.53 \\
    General Rubric   & 49.00 & 86.05 & \textbf{34.76} & 61.90 & 45.97 & \textbf{39.19} & 45.66 & 51.79 \\
    Reference Answer & 49.01 & 85.67 & 29.88 & 61.40 & \textbf{50.13} & 37.12 & 10.54 & 46.25 \\
    \midrule
    ARES-RL          & \textbf{49.36} & \textbf{86.96} & \textbf{34.76} & \textbf{63.16} & 41.45 & 38.24 & \textbf{54.88} & \textbf{52.69} \\
    \bottomrule
    \end{tabular}
    }
    \vspace{-15pt}
    \end{table}


Table~\ref{tab:ablation} reports the ablation results. 
We emphasize that no reward strategy dominates every individual benchmark. 
Different reward designs encode different inductive biases: reference-based rewards can be strong when benchmark evaluation aligns with semantic similarity or content coverage relative to a canonical answer, while general rubrics can be competitive when broad quality dimensions are sufficient. 
Therefore, we interpret the ablation primarily in terms of cross-benchmark robustness rather than per-benchmark dominance. 
Under this view, ARES-RL achieves the best average score and is top-performing or tied on five out of seven benchmarks, while avoiding the severe instability observed in reference-based evaluation.

\vspace{-5pt}
\paragraph{Structured Rubrics Outperform Holistic Judgments on Average.}
Both rubric-based approaches, ARES-RL (52.69) and General Rubric (51.79), outperform the Blind Judge baseline (49.53) on average. 
This indicates that decomposing evaluation into explicit criteria provides a more effective reward signal than asking an LLM judge to produce a single holistic quality score. 
Rather than relying on an undifferentiated judgment, rubric evaluation makes the reward decision more transparent and provides more actionable feedback for policy optimization. 
The improvement of General Rubric over Blind Judge also suggests that even simple structured criteria can help stabilize reward estimation.

\vspace{-5pt}
\paragraph{Question-Specific Rubrics Improve Cross-Task Robustness.}
ARES-RL outperforms the General Rubric baseline by 0.90 points on average, with the largest gain on IFEval (+9.22). 
However, question-specific rubrics do not uniformly improve every benchmark: General Rubric performs better on HealthBench and WritingBench, suggesting that generic quality dimensions can already be effective for some broad evaluation settings. 
The advantage of ARES-RL is most evident on tasks whose evaluation requirements vary substantially across examples. 
For instance, instruction-following tasks may require different combinations of formatting, length, keyword, language, and structural constraints, which are difficult to express with a fixed set of generic criteria such as accuracy or clarity. 
Question-specific rubrics address this limitation by aligning the reward criteria with each generated question and reference answer, leading to better average performance and stronger robustness.

\paragraph{Reference-Based Evaluation Is Strong but Unstable.}
The Reference Answer reward achieves the highest HealthBench score (50.13), suggesting that reference-based comparison can be effective when the benchmark favors content coverage close to a canonical answer. 
This is plausible for health-related QA, where high-quality responses often require specific factual content, risk disclosure, and cautious recommendations that may be well captured by a strong reference answer. 
However, this reward design is much less stable across task types: it drops sharply on IFEval (10.54), where success depends less on matching a single target answer and more on satisfying explicit constraints such as formatting, length, keywords, and structural requirements. 
Consequently, Reference Answer obtains the lowest average score despite its strong HealthBench performance. 
This highlights the limitation of using a single reference answer as the reward target for heterogeneous open-ended RL training.

\begin{figure*}[t]
    \centering
    \includegraphics[width=0.9\textwidth]{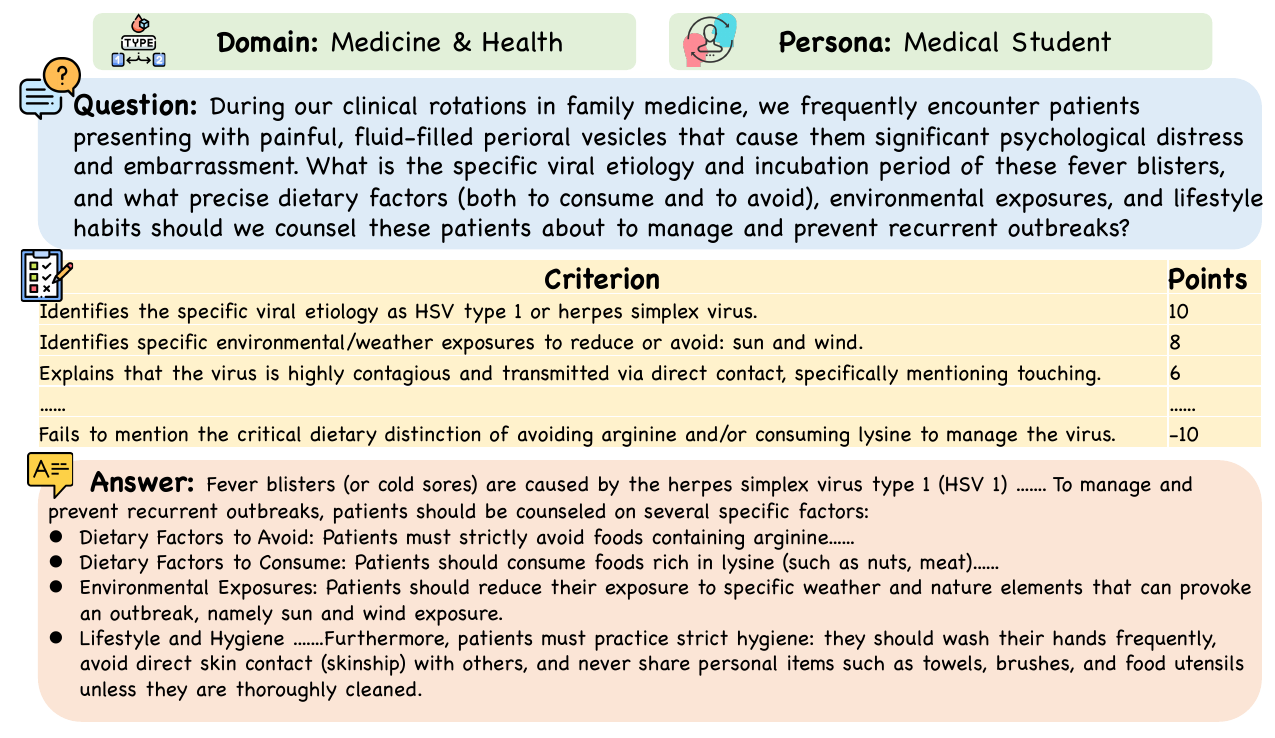}
    \vspace{-1em}
    \caption{A representative ARES-generated instance from the Medicine \& Health domain.}
    \label{fig:case_study}
    \vspace{-1em}
\end{figure*}

\subsection{Case Study}
Figure~\ref{fig:case_study} shows a representative ARES instance from the medicine domain. The rubric decomposes an open-ended clinical question into weighted criteria: positive weights reward key knowledge dimensions (e.g., HSV-1 identification, +10; environmental triggers, +8), while negative weights penalize critical omissions (e.g., missing the arginine/lysine distinction, $-$10). Unlike a binary reward that would assign zero to any incomplete response, this structure provides partial credit proportional to the importance of satisfied criteria, guiding the policy toward more comprehensive outputs.

\section{Conclusion}
\label{sec:conclusion}

We presented ARES, a scalable pipeline that automatically converts raw pretraining documents into RL-ready instances with question-specific weighted rubrics. By co-generating questions, reference answers, and multi-dimensional rubrics in a single pass, ARES produces 100K rubric-annotated instances across ten domains at negligible additional cost. Experiments on seven benchmarks show that ARES-RL consistently outperforms CPT, SFT, and binary-reward RL, with the largest gains on open-ended tasks where multi-dimensional rubric rewards provide structured supervision that binary verification cannot offer.

\bibliographystyle{plainnat}
\bibliography{references}

\appendix

\section{Limitations}
\label{sec:limitations}

Due to computational resource constraints, our experiments are conducted exclusively on Qwen3-4B-Base. This limits the scope of our findings. We are unable to validate whether ARES scales to larger models (e.g., 70B parameters), where rubric-based rewards may yield different dynamics due to increased model capacity and stronger baseline reasoning. Second, the rubric judge (Qwen3-32B) is constrained by our available GPU memory; using a larger or more capable judge model could improve reward signal quality but is infeasible within our current setup. We believe these are engineering rather than fundamental limitations, and we expect the core findings---that question-specific rubric rewards outperform binary and holistic alternatives---to hold at larger scales.

\section{Broader Impact}
\label{sec:broader_impact}

ARES enables RL training on open-ended tasks that were previously inaccessible to verifiable-reward methods, broadening LLM capabilities in domains such as healthcare, education, and professional writing. On the positive side, by automating rubric synthesis from pretraining documents, ARES lowers the barrier to constructing diverse RL training data, making rubric-based post-training accessible to research groups with limited annotation budgets. However, this broader applicability also carries risks. Automated rubric generation may encode biases present in the source documents or the LLMs used for synthesis, potentially reinforcing stereotypes or inaccuracies. The use of LLM judges for reward assignment introduces a further layer of model-dependent bias. We encourage practitioners to carefully validate ARES-generated data and rubrics, and to combine automated rubric evaluation with human oversight when deploying models trained with this pipeline.

\section{Additional Implementation Details}
\label{app:implementation}

This section provides additional details on the ARES data construction and
training setup. The main paper is self-contained; the material below is intended
to clarify implementation choices and reproducibility details.

\subsection{Pipeline Inputs and Outputs}
\label{app:pipeline_io}

ARES takes raw pretraining documents as input and produces GRPO-ready training
instances. Each accepted instance contains a user-facing prompt, an
instance-specific rubric, a reference answer, and metadata describing the source
domain and persona. In the training format, the prompt is represented as a
single-turn conversation, the rubric is stored as a list of natural-language
criteria with non-zero integer weights, and the reference answer is retained as
an ideal completion for analysis and supervised baselines.
The metadata is not used as a reward during RL training. It is retained for
auditing, filtering, and dataset analysis.

\subsection{Domain and Persona Assignment}
\label{app:domain_persona}

The domain classifier assigns each document to one of ten categories:
Math, Coding, Medicine \& Health, Technology \& Engineering, Social
Science, Natural Science, Travel \& Lifestyle, Commerce \& Economics, Education,
and Other. If a document plausibly spans multiple categories, the classifier
selects the most relevant primary domain. Persona conditioning is used to
diversify the resulting questions: each document can be paired with up to three
audience profiles, and the generator produces questions from the selected
persona's perspective.

This design separates topical coverage from user intent. For example, two
documents in the same medical domain can yield different questions when framed
for a medical professional, a student, or a caregiver. Conversely, documents from
different domains can share a similar persona while still requiring different
rubric criteria.

\section{Rubric Generation and Validation}
\label{app:rubric_details}

\subsection{Rubric Format}
\label{app:rubric_format}

Each rubric consists of a list of criteria
$\mathcal{R}=\{(c_i,w_i)\}_{i=1}^{N}$, where $c_i$ is a natural-language
criterion and $w_i$ is a non-zero integer weight. Positive weights reward
desirable properties of an answer, while negative weights penalize important
failure modes. Rubrics are generated jointly with the question and reference
answer so that all three objects are grounded in the same source document.

The generator is instructed to write semantic criteria rather than string-match
rules. A criterion should describe the concept, fact, reasoning step, constraint,
or failure mode to be checked; it should not require reproducing a particular
surface form unless the task itself is about formatting or exact wording.

\subsection{Complexity-Adaptive Criteria}
\label{app:criteria_count}

The number of criteria is adapted to question complexity. Easy questions use a
small number of criteria focused on the central fact or concept. Medium questions
use more criteria to cover multiple related aspects. Hard questions use the most
criteria, including separate checks for sub-conditions, exceptions, and common
failure modes. Across all complexity levels, the pipeline requires both positive
and negative criteria so that the reward can distinguish partially correct
answers from answers that omit or distort critical information.

\subsection{Validation Checks}
\label{app:validation}

After generation, each rubric is validated before being serialized for training.
The validator checks that every criterion has the required fields, that weights
are non-zero and within the allowed range, that the rubric contains enough
criteria for the corresponding complexity level, and that the rubric includes
negative criteria. Criteria with malformed fields are rejected or repaired when
the correction is unambiguous; examples that remain structurally invalid are not
used for training.

The verifier also checks the associated Q\&A pair before rubric validation. A
generated instance must pass factual consistency, information-leakage, and
self-containedness checks. This prevents the RL stage from receiving prompts
whose answers are unsupported by the source document, trivially revealed by the
question, or impossible to interpret without hidden context.

\section{Training Details}
\label{app:training_details}

\subsection{Compared Methods}
\label{app:compared_methods}

All methods in the main comparison use Qwen3-4B-Base as the foundation model.
Continual pretraining optimizes next-token likelihood on the same source
documents used by ARES. Natural Reasoning fine-tunes on an off-the-shelf
reasoning dataset. Webscale follows binary-reward RL on short-answer QA pairs.
ARES-SFT fine-tunes on the generated ARES questions and reference answers.
ARES-RL uses the same generated questions but optimizes rubric-based rewards.

This comparison is designed to isolate three factors: whether the model sees the
source documents as language modeling data or as question-answer tasks, whether
the training signal is imitation or reinforcement learning, and whether the
reward is binary or multi-dimensional.

\subsection{GRPO Configuration}
\label{app:grpo_config}

ARES-RL uses GRPO with group-normalized advantages. For each prompt, the policy
samples a group of candidate responses, scores each response with the associated
rubric, and updates the policy using the clipped GRPO objective described in the
main paper. The experiments use $G=8$ rollouts per question, clipping threshold
$\varepsilon=0.2$, KL penalty $\beta=0.01$, batch size 128, and 3 epochs with VeRL as the training backend.

The rubric score used for RL is the weighted sum of binary criterion judgments.
The judge evaluates each criterion independently against the sampled answer, and
the resulting weighted score is used as the scalar reward for GRPO.

\subsection{SFT and CPT Hyperparameters}
\label{app:hyperparams}

Table~\ref{tab:hyperparams} lists the training hyperparameters for the NaturalReasoning-SFT/ARES-SFT and CPT settings. 

\begin{table}[t]
  \centering
  \caption{Training hyperparameters for NaturalReasoning-SFT/ARES-SFT and CPT baselines.}
  \label{tab:hyperparams}
  \small
  \begin{tabular}{lcc}
  \toprule
  Setting & NaturalReasoning-SFT / ARES-SFT & CPT \\
  \midrule
  Base model          & Qwen3-4B-Base & Qwen3-4B-Base \\
  Data mode           & prompt + response & text \\
  GPUs                & 8 & 8 \\
  Max length          & 2048 & 2048 \\
  Micro batch / GPU   & 2 & 4 \\
  Global batch size   & 128 & 256 \\
  Epochs              & 3 & 3 \\
  Learning rate       & 2e-5 & 5e-5 \\
  Warmup ratio        & 0.05 & 0.1 \\
  Scheduler           & cosine & cosine \\
  Betas               & [0.9, 0.95] & [0.9, 0.95] \\
  Weight decay        & 0.01 & 0.01 \\
  Grad clip           & 1.0 & 1.0 \\
  Strategy            & FSDP2 & FSDP2 \\
  Gradient ckpt       & enabled & enabled \\
  LoRA                & disabled & disabled \\
  Seed                & 42 & 42 \\
  \bottomrule
  \end{tabular}
\end{table}

\subsection{Compute Resources}
\label{app:compute}

All training experiments are conducted on a single node equipped with 8$\times$ NVIDIA A100 80GB GPUs, 224 CPU cores, and 2048 GB of system memory. The total compute budget across all training runs amounts to approximately 140 GPU-hours. 

\section{Evaluation Details}
\label{app:evaluation_details}

The evaluation suite covers seven benchmarks: MMLU-Pro, GSM8K, HumanEval+,
MBPP+, HealthBench, WritingBench, and IFEval. Structured tasks use their standard
automatic evaluation protocols, including exact or normalized answer matching
for math and test-case execution for code. Open-ended tasks are evaluated with
their benchmark-specific rubric or judge protocols, which makes them especially
relevant for testing whether rubric-based RL improves multi-dimensional response
quality.

We report the unweighted average over the seven benchmark scores in the main
tables. This average is intended as a compact summary of broad capability rather
than a claim that all benchmarks have identical practical importance.

To quantify evaluation uncertainty without changing the reported point
estimates, we compute 95\% confidence intervals. The intervals are shown in Tables~\ref{tab:main_ci_struct}--\ref{tab:main_ci_open} and Tables~\ref{tab:ablation_ci_struct}--\ref{tab:ablation_ci_open}.

\begin{table}[t]
    \centering
    \caption{Main benchmark results with 95\% confidence intervals (structured tasks).}
    \label{tab:main_ci_struct}
    \resizebox{\textwidth}{!}{
    \begin{tabular}{lcccc}
    \toprule
    Method & MMLU-Pro & GSM8K & HumanEval & MBPP \\
    \midrule
    CPT & 46.02 [45.11, 46.92] & 82.34 [80.29, 84.39] & 32.32 [25.03, 39.66] & 59.40 [54.63, 64.15] \\
    Natural Reasoning & 47.96 [47.06, 48.86] & 81.50 [79.46, 83.55] & 32.93 [25.67, 40.30] & 61.15 [56.36, 65.88] \\
    Webscale & 49.50 [48.62, 50.38] & 84.91 [82.95, 86.81] & 33.54 [26.23, 40.86] & 61.40 [56.64, 66.16] \\
    ARES-SFT & 50.56 [49.67, 51.46] & 85.67 [83.69, 87.55] & 31.10 [23.82, 38.46] & 61.40 [56.63, 66.16] \\
    ARES-RL (ours) & 49.36 [48.48, 50.25] & 86.96 [85.13, 88.77] & 34.76 [27.45, 42.08] & 63.16 [58.37, 67.89] \\
    \bottomrule
    \end{tabular}}
\end{table}

\begin{table}[t]
    \centering
    \caption{Main benchmark results with 95\% confidence intervals (open-ended tasks).}
    \label{tab:main_ci_open}
    \resizebox{\textwidth}{!}{
    \begin{tabular}{lcccc}
    \toprule
    Method & HealthBench & WritingBench & IFEval & Avg \\
    \midrule
    CPT & 35.04 [34.07, 36.04] & 36.98 [36.08, 37.88] & 39.39 [36.02, 42.77] & 47.36 [46.01, 48.76] \\
    Natural Reasoning & 32.94 [32.01, 33.88] & 36.77 [36.05, 37.50] & 28.15 [25.06, 31.30] & 45.91 [44.55, 47.28] \\
    Webscale & 36.08 [35.12, 37.03] & 37.09 [36.20, 38.01] & 35.61 [32.38, 38.87] & 48.30 [46.94, 49.66] \\
    ARES-SFT & 35.78 [34.77, 36.81] & 37.05 [36.12, 38.02] & 46.41 [42.94, 49.93] & 49.71 [48.34, 51.08] \\
    ARES-RL (ours) & 41.45 [40.34, 42.56] & 38.24 [37.31, 39.21] & 54.88 [51.34, 58.31] & 52.69 [51.31, 54.07] \\
    \bottomrule
    \end{tabular}}
\end{table}

\begin{table}[t]
    \centering
    \caption{Reward-design ablation results with 95\% confidence intervals (structured tasks).}
    \label{tab:ablation_ci_struct}
    \resizebox{\textwidth}{!}{
    \begin{tabular}{lcccc}
    \toprule
    Reward Strategy & MMLU-Pro & GSM8K & HumanEval & MBPP \\
    \midrule
    Blind Judge & 47.56 [46.67, 48.45] & 85.97 [84.08, 87.80] & 33.54 [26.76, 40.79] & 62.16 [57.35, 66.87] \\
    General Rubric & 49.00 [48.12, 49.88] & 86.05 [84.16, 87.87] & 34.76 [27.50, 42.13] & 61.90 [57.16, 66.68] \\
    Reference Answer & 49.01 [48.11, 49.91] & 85.67 [83.77, 87.56] & 29.88 [23.17, 37.19] & 61.40 [56.68, 65.95] \\
    ARES-RL & 49.36 [48.46, 50.23] & 86.96 [85.15, 88.71] & 34.76 [27.46, 42.09] & 63.16 [58.37, 67.90] \\
    \bottomrule
    \end{tabular}}
\end{table}

\begin{table}[t]
    \centering
    \caption{Reward-design ablation results with 95\% confidence intervals (open-ended tasks).}
    \label{tab:ablation_ci_open}
    \resizebox{\textwidth}{!}{
    \begin{tabular}{lcccc}
    \toprule
    Reward Strategy & HealthBench & WritingBench & IFEval & Avg \\
    \midrule
    Blind Judge & 44.43 [43.25, 45.59] & 38.84 [37.94, 39.77] & 34.20 [30.81, 37.69] & 49.53 [48.15, 50.91] \\
    General Rubric & 45.97 [44.84, 47.14] & 39.19 [38.22, 40.16] & 45.66 [42.13, 49.11] & 51.79 [50.39, 53.21] \\
    Reference Answer & 50.13 [48.86, 51.42] & 37.12 [36.33, 37.93] & 10.54 [8.22, 12.90] & 46.25 [44.97, 47.57] \\
    ARES-RL & 41.45 [40.34, 42.55] & 38.24 [37.31, 39.19] & 54.88 [51.49, 58.32] & 52.69 [51.32, 54.04] \\
    \bottomrule
    \end{tabular}}
\end{table}

\section{Prompt Templates}
\label{app:prompts}

We include representative prompt excerpts used by the ARES data pipeline. The prompts below are lightly compressed for readability while preserving the operational instructions used in each stage.

\subsection{Document Filtering Prompt}
\label{app:filter_prompt}

\begin{promptbox}[Stage 1: Document Filtering]
You are a data quality analyst. You will be given material from diverse sources. Our goal is to generate question-answer pairs WITH evaluation rubrics from the material.

Identify whether the material is qualified for BOTH Q\&A generation AND rubric creation based on these criteria:

\textbf{Content Quality:}
\begin{itemize}[leftmargin=*, itemsep=0pt, topsep=2pt]
  \item The material is informative and factually accurate.
  \item It contains specific, verifiable information.
  \item It is self-contained enough to formulate clear questions.
\end{itemize}

\textbf{Suitable for Rubric Generation:}
\begin{itemize}[leftmargin=*, itemsep=0pt, topsep=2pt]
  \item It contains concrete facts or concepts that can be evaluated objectively.
  \item It allows distinguishing between correct and incorrect answers.
  \item It has sufficient detail to create meaningful evaluation criteria.
  \item It provides clear standards such as numbers, dates, definitions, or procedures.
\end{itemize}

\textbf{Unsuitable Materials:} Pure opinion pieces without factual basis; overly vague or ambiguous content; lists without context; promotional content without informative value; content with unclear or disputed facts.

\textbf{Output JSON:} \{"thought": "...", "qualified": "Y or N"\}
\end{promptbox}

\subsection{Domain and Persona Identification Prompt}
\label{app:identifier_prompt}

\begin{promptbox}[Stage 2: Domain \& Persona Identification]
You are a helpful data analyst. You will be given material from a website which can come from very diverse sources and may not be well structured. Our final goal is to generate a question-answer pair from the material. In this stage, identify the domain and persona of the material.

\textbf{Instructions:}
\begin{itemize}[leftmargin=*, itemsep=0pt, topsep=2pt]
  \item Choose one domain from: Math, Technology \& Engineering, Coding, Social Science, Natural Science, Travel \& Lifestyle, Commerce \& Economics, Medicine \& Health, Education, Other.
  \item If multiple domains apply, choose the most relevant primary domain.
  \item The persona is the intended audience of the material.
  \item If multiple personas apply, list up to three personas.
\end{itemize}

\textbf{Output JSON:} \{"thought": "...", "domain": "...", "persona": "..."\}
\end{promptbox}

\subsection{Rubric-Augmented QA Generation Prompt}
\label{app:generator_prompt}

\begin{promptbox}[Stage 3: Rubric-Augmented QA Co-Generation]
You will be given material from diverse sources. Your task is to: (1) Generate a question-answer pair from the material. (2) Generate evaluation rubrics based on question complexity.

\textbf{Question \& Answer Generation:}
\begin{itemize}[leftmargin=*, itemsep=0pt, topsep=2pt]
  \item Generate from the persona's perspective.
  \item Base the question and answer ONLY on the material content.
  \item The question must be clear and verifiable.
  \item The answer should be comprehensive based on the material depth.
  \item Add necessary background context because the material will not be provided during training.
  \item Never leak the answer in the question.
\end{itemize}

\textbf{Evaluation Rubrics:}
\begin{itemize}[leftmargin=*, itemsep=0pt, topsep=2pt]
  \item Simple questions: 4--6 criteria for a single fact or concept.
  \item Medium questions: 6--10 criteria for multiple related points.
  \item Complex questions: 10--15 criteria for multifaceted answers with conditions, exceptions, or failure modes.
\end{itemize}

\textbf{Scoring:}
\begin{itemize}[leftmargin=*, itemsep=0pt, topsep=2pt]
  \item Positive weights reward desirable answer properties.
  \item Negative weights penalize critical omissions, factual errors, or misleading statements.
  \item Each criterion should check one specific aspect.
  \item Scores should reflect the importance of that aspect in context.
\end{itemize}

\textbf{Semantic Rubric Guidelines:} Reference concrete concepts, facts, numbers, terms, or constraints from the material. Do not rely on exact string matching unless the task itself requires exact wording. Use negative criteria for common errors or important missing information.

\textbf{Output JSON:} \{"thought": "...", "question": "...", "answer": "...", "rubrics": [\{"criterion": "...", "points": 9\}, \{"criterion": "...", "points": 7\}, \{"criterion": "...", "points": -8\}]\}
\end{promptbox}

\subsection{Q\&A and Rubric Suitability Checking Prompt}
\label{app:checker_prompt}

\begin{promptbox}[Stage 4--5: Quality \& Rubric Validation]
You are a data quality checker for Q\&A pairs that will be used to generate evaluation rubrics. Check the Q\&A pair quality and assess its suitability for rubric generation.

\textbf{Basic Q\&A Checks:}
\begin{itemize}[leftmargin=*, itemsep=0pt, topsep=2pt]
  \item Context sufficiency: the question should be self-contained and should not require seeing the original material.
  \item Answer correctness: the answer must be factually correct according to the material and must not add unsupported information.
  \item Information leakage: the question should not explicitly reveal the answer.
\end{itemize}

\textbf{Rubric Suitability:}
\begin{itemize}[leftmargin=*, itemsep=0pt, topsep=2pt]
  \item Determine whether the answer contains concrete, verifiable facts.
  \item Determine whether multiple aspects can be objectively evaluated.
  \item Determine whether clear standards exist for good and bad responses.
  \item Reject examples that are too short, purely subjective, vague, yes/no only, or lack specific details.
\end{itemize}

\textbf{Output JSON:} \{"thought": "...", "has\_context": "Y/N", "answer\_correctness": "Y/N", "info\_leakage": "Y/N", "question\_complexity": "simple/medium/complex", "answer\_detail\_level": "low/medium/high", "rubric\_suitable": "Y/N"\}
\end{promptbox}

\section{Licenses for Existing Assets}
\label{app:licenses}

We list the licenses of all major external assets used in this work. All assets are used in compliance with their respective terms.

\begin{table}[h]
    \centering
    \small
    \caption{Licenses for existing assets used in this work.}
    \label{tab:licenses}
    \begin{tabular}{lll}
    \toprule
    \textbf{Asset} & \textbf{Type} & \textbf{License} \\
    \midrule
    \multicolumn{3}{l}{\emph{Pretraining Corpora}} \\
    DCLM~\citep{li2024datacomp}         & Dataset & MIT \\
    FineWeb-Edu~\citep{penedo2024fineweb} & Dataset & ODC-By-1.0 \\
    FinePDFs~\citep{finepdfs2024}       & Dataset & ODC-By-1.0 \\
    \midrule
    \multicolumn{3}{l}{\emph{Models}} \\
    Qwen3-4B-Base~\citep{yang2024qwen2} & Model & Apache-2.0 \\
    Qwen3-32B~\citep{yang2024qwen2}     & Model & Apache-2.0 \\
    \midrule
    \multicolumn{3}{l}{\emph{Training Framework}} \\
    VeRL~\citep{sheng2025hybridflow}    & Code & Apache-2.0 \\
    \midrule
    \multicolumn{3}{l}{\emph{Evaluation Benchmarks}} \\
    MMLU-Pro~\citep{wang2024mmlu}       & Benchmark & MIT \\
    GSM8K~\citep{cobbe2021gsm8k}        & Benchmark & MIT \\
    HumanEval+~\citep{chen2021humaneval} & Benchmark & Apache-2.0 \\
    MBPP+~\citep{austin2021mbpp}        & Benchmark & Apache-2.0 \\
    HealthBench~\citep{healthbench2025} & Benchmark & MIT \\
    WritingBench~\citep{writingbench2025} & Benchmark & Apache-2.0 \\
    IFEval~\citep{zhou2023ifeval}       & Benchmark & Apache-2.0 \\
    \midrule
    \multicolumn{3}{l}{\emph{Baseline Datasets}} \\
    NaturalReasoning~\citep{yuan2025naturalreasoning} & Dataset & CC-BY-NC-4.0 \\
    Webscale-RL~\citep{cen2025webscale} & Dataset & CC-BY-NC-4.0 \\
    \bottomrule
    \end{tabular}
\end{table}

\section{Use of Large Language Models}
\label{app:llm_usage}

This work uses large language models in two capacities.
First, as described in the main paper, LLMs serve as core components of the ARES pipeline: they perform document filtering, domain and persona identification, rubric-augmented question--answer generation, quality validation, and rubric-based reward judging during RL training.
These uses are detailed in the method and experiment sections.
Second, we used LLM-based assistants to polish the writing of this manuscript, including grammar correction, sentence rephrasing, and stylistic refinement.
All LLM-generated or LLM-edited text has been manually reviewed and revised by the authors.
We take full responsibility for the accuracy and integrity of every claim, result, and statement in this paper.

\end{document}